\documentclass[10pt,twocolumn,letterpaper]{article}

\usepackage{wacv}
\usepackage{times}
\usepackage{epsfig}
\usepackage{graphicx}
\usepackage{amsmath}
\usepackage{amssymb}
\usepackage{booktabs}
\usepackage{caption}
\usepackage{tabularx}
\newcommand\setrow[1]{\gdef\rowmac{#1}#1\ignorespaces}
\newcommand\clearrow{\global\let\rowmac\relax}
\clearrow
\def\wacvPaperID{881} % Enter the WACV Paper ID here

\wacvapplicationstrack % Uncomment this line if you are submitting to the Applications Track.

\wacvfinalcopy % *** Uncomment this line for the final submission

\ifwacvfinal
\usepackage[breaklinks=true,bookmarks=false]{hyperref}
\else
\usepackage[pagebackref=true,breaklinks=true,colorlinks,bookmarks=false]{hyperref}
\fi

\begin{document}

%%%%%%%%% TITLE
\title{Color Recommendation for Vector Graphic Documents based on Multi-Palette Representation}

\author{Qianru Qiu\\
CyberAgent\\
Shibuya, Tokyo, Japan\\
{\tt\small qiu\_qianru@cyberagent.co.jp}
\and
Xueting Wang\\
CyberAgent\\
Shibuya, Tokyo, Japan\\
{\tt\small wang\_xueting@cyberagent.co.jp}
\and
Mayu Otani\\
CyberAgent\\
Shibuya, Tokyo, Japan\\
{\tt\small otani\_mayu@cyberagent.co.jp}
\and
Yuki Iwazaki\\
CyberAgent\\
Shibuya, Tokyo, Japan\\
{\tt\small iwazaki\_yuki@cyberagent.co.jp}
}

\twocolumn[{
\maketitle
\begin{center}
    \captionsetup{type=figure}
    \includegraphics[width=1.0\linewidth]{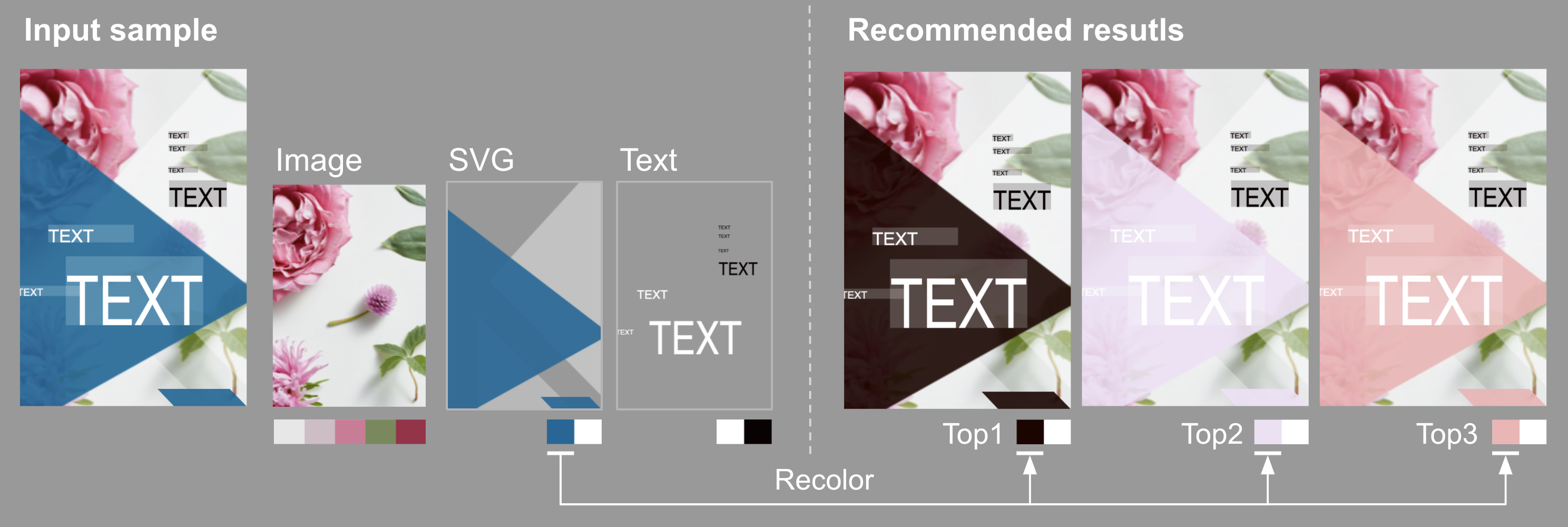}
    \captionof{figure}{Color recommendation for vector graphic documents. On the left are the design sample and its visual elements (e.g. Image, SVG and Text) with extracted palettes. On the right are the recommended top3 results recoloring one SVG color. The elements in the input sample are from the Crello dataset.}
    \label{fig:overview}
\end{center}
}]

% \maketitle
\thispagestyle{empty}

%%%%%%%%% ABSTRACT
\begin{abstract}
   Vector graphic documents present multiple visual elements, such as images, shapes, and texts. Choosing appropriate colors for multiple visual elements is a difficult but crucial task for both amateurs and professional designers. Instead of creating a single color palette for all elements, we extract multiple color palettes from each visual element in a graphic document, and then combine them into a color sequence. We propose a masked color model for color sequence completion and recommend the specified colors based on color context in multi-palette with high probability. We train the model and build a color recommendation system on a large-scale dataset of vector graphic documents. The proposed color recommendation method outperformed other state-of-the-art methods by both quantitative and qualitative evaluations on color prediction and our color recommendation system received positive feedback from professional designers in an interview study.
\end{abstract}

%%%%%%%%% BODY TEXT
\section{Introduction}

In graphic design, there are many creative applications providing thousands of templates. These design platforms are suitable for creative designers and amateurs such as marketing professionals, bloggers, social media managers, etc. In design workflows, users choose a template and replace the elements using their own assets. The pre-designed templates have coordinated colors in each visual element. When some visual elements are replaced, the color harmony may be destroyed. Selecting appropriate colors is not easy for amateurs, even designers usually struggle with getting suitable color palettes for vector graphic documents.

A color palette refers to a limited number of colors expressed in refined forms. It is widely used in graphic design due to its simplicity, intuitiveness, generality, and easy computation \cite{kim2021dynamic}. The prior researches of color palette representation proposed to train a regression model \cite{o2011color, kita2016aesthetic}. These regression methods extract hundreds of color features manually and learn the weights of each feature. The feature extraction is complex including palette colors, mean, standard deviation, median, max, min, and max minus min across a single channel in each color space, i.e., RGB, CIELAB, and HSV. The difficulty of the hand-crafted features is that they do not comprehensively encode the semantics of colors, and some features might not have significant effects in terms of the downstream tasks. Learning high-quality representation of color remains an open problem.
In this work, we simplify the input without hand-crafted features and propose a data-driven deep learning model for color representation.
% We explore harmonious color prediction in the similar way of word embedding methods, where the input is a color corpus and the output is a set of feature vectors. 
% We propose to train a color model with the input of color code and apply word embedding methods to learn color distributed representation from the color palettes.

In recent years, some researchers have explored deep learning techniques on color palette generation and color recommendation. The previous researches focus on generating a color palette for a single visual target, such as image colorization \cite{bahng2018coloring}, shapes colorization in statistical graphics \cite{lu2020palettailor}, shapes and texts colorization in infographics \cite{yuan2021infocolorizer}. However, a vector graphic document is much more complex with multiple visual elements, including images, shapes, and texts simultaneously. Each visual element has its own palette. It is challenging for existing color recommendation tools to recommend colors for multi-palette design.
In this study, we use a color sequence combining multiple palettes of different visual elements and train a masked color model to learn multi-palette representation by color sequence completion.

In summary, our main contributions include:
\begin{itemize}
    \setlength{\itemsep}{-5pt}
    \item A novel masked color model to represent multiple palettes in vector graphic documents.
    \item An interactive system for color recommendation which recolors graphic documents with the recommended colors.
    \item A series of experimental evaluations covering quantitative experiments and perceptual studies for the recommendation system and the recommended results that validates the effectiveness of the proposed methods.
\end{itemize}

\section{Related works}
\subsection{Color recommendation}

There are mainly two scenarios of color recommendation. The first one is to suggest a color palette for specified themes or semantic requirements. The second one is to expand a color palette based on the given colors. For the first scenario, there are some websites, such as Adobe Color \cite{AdobeColor}, and COLOURLovers \cite{ColourLovers}, providing color palette templates classified with various theme names or semantic tags, such as 'natural', 'environmental'. These palette templates can be used as references to suit semantic requirements. Some researches suggest color palette templates based on semantic tags and fixed harmonic color selection models to generate text colors based on image colors in magazine cover design \cite{jahanian2013recommendation, yang2016automatic}. 
% The previous color harmony models evaluate two or three color combinations in color theory. 
For the second scenario, an early effort by O’Donovan \etal. \cite{o2011color} proposed a linear regression method and suggested the fifth color for the given four colors. Kita \etal. \cite{kita2016aesthetic} used the same regression method to expand a color palette composed of N colors to $N+\alpha$ retraining the original color harmony. These regression models depended on the hand-crafted feature extraction methods. In this work, we suggest compatible colors for the given colors in multi-palette without color feature extraction methods by a deep learning model. 

Recently, some researchers have explored deep learning algorithms on color palette recommendation. Yuan \etal. \cite{yuan2021infocolorizer} employed a Variational AutoEncoder with Arbitrary Conditioning (VAEAC) model to generate a color palette for infographic elements. Kim \etal. \cite{kim2022colorbo} trained the color embedding model to predict and recommend other colors that are likely to gather together in the same palette for Mandala Coloring. The color model was in a similar way as fastText \cite{bojanowski2017enriching}, an extension of Word2Vec \cite{mikolov2013efficient} model. They signified a color as a word, and the palette as a sentence. The model was trained providing a continuous vector representation of colors. The elements in these design objects include shapes, or texts, and each element is limited to a single color. In graphic documents, the elements also include photos and illustrations. The color design in vector graphic documents is more complex as a multi-palette design. 
We extend the idea of applying word embedding to color representation similar to the previous Word2Vec-based work \cite{kim2022colorbo}. However, the Word2Vec models aren't able to account for the relationships between different palettes in the same graphic document. We explore multi-palette representation by contextual embedding based on BERT architecture \cite{kenton2019bert}. 

\subsection{Palette-based image recoloring}

There are some image recoloring approaches based on semantic segmentation from image by deep neural networks \cite{afifi2019image, khodadadeh2021automatic}. We focus on palette-based models that a color palette is useful and easy to express the color concept of an image.
Color palette captures the main colors in an image and is used to adjust the color composition of an image towards the desired color palette. Most approaches for recoloring images involve two steps, as extracting a palette from the image and mapping every pixel in the image to the target palette. Many approaches \cite{chang2015palette, zhang2017palette, akimoto2020fast} employ clustering methods to extract palette colors. Several other works \cite{tan2018efficient, wang2019improved} use a geometric method to extract palettes that construct a convex hull in RGB color space. The convex-hull-based palettes may miss important colors that lie within the convex hull. We use k-means clustering method to recolor image elements in our system.

\section{Multi-palette representation}

\subsection{Datasets}

\begin{figure}[h]
\begin{center}
   \includegraphics[width=1.0\linewidth]{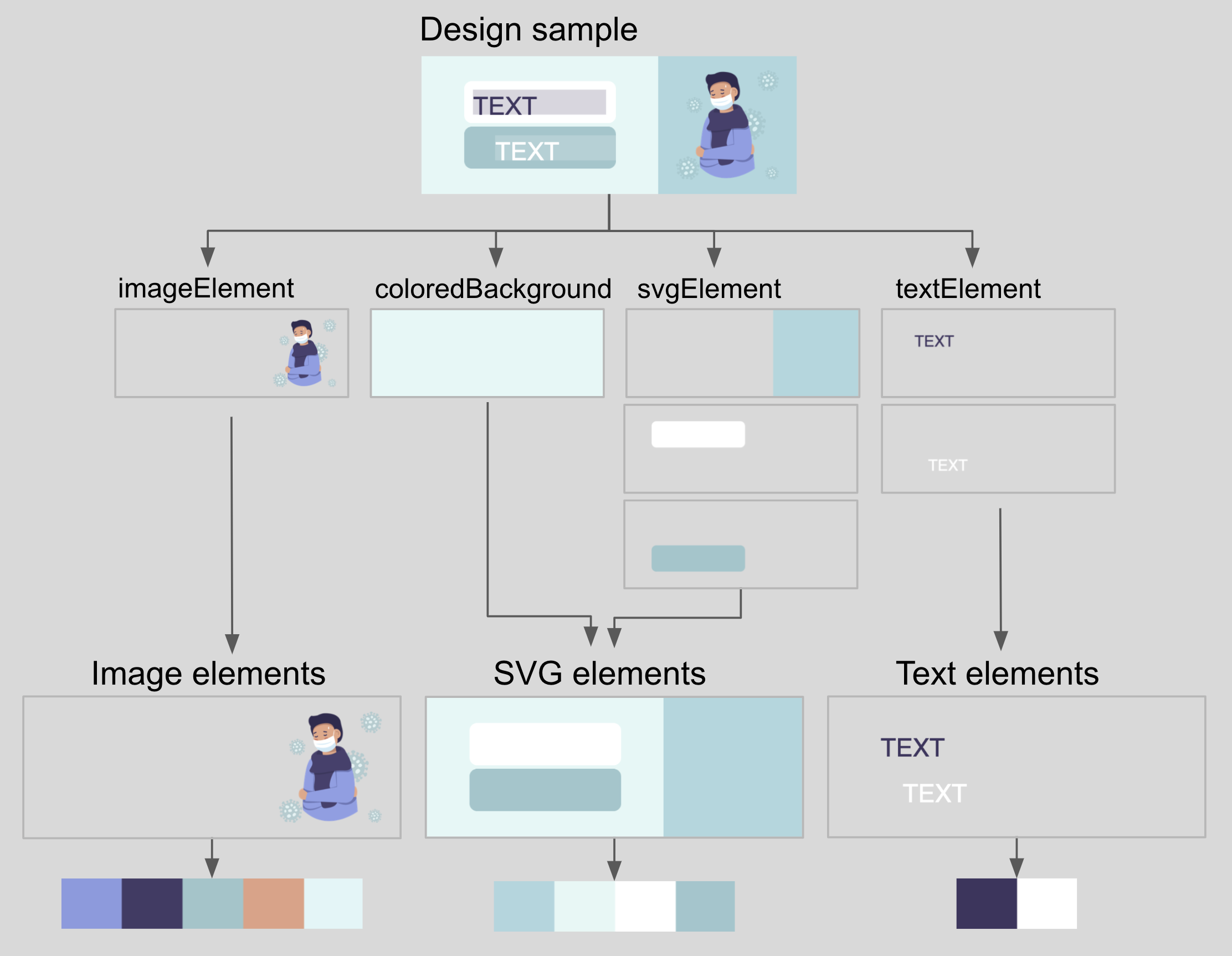}
\end{center}
   \caption{Multi-palette extraction from a design template as Image-SVG-Text palettes. Merge the elements in the same element group into a single image and then extract the color palette.}
\label{fig:paletteExt}
\end{figure}

We generate a multi-palette dataset from a large-scale dataset Crello \cite{yamaguchi2021canvasvae}. The Crello dataset contains design templates for various display formats, such as social media posts, banner ads, blog headers, or printed posters. It offers complete document structure and element attributes including element-specific configuration, such as type of the element, position, size, opacity, color, or a raster image. The element types mainly include imageElement, maskElement, coloredBackground, svgElement, and textElement. We classify the elements into three groups, as image element group including imageElement and maskElement, scalable vector graphic (SVG) element group including coloredBackground and svgElement, and text element group including textElement. The color data of each element in the Crello dataset only has 1 color that is relevant for solid background and text placeholder. We generate a multi-palette dataset as Image-SVG-Text palettes that each element group has its own palette as shown in Figure~\ref{fig:paletteExt}. For image and SVG elements, we merge the elements of the same group into a single image and then extract the color palette using k-means clustering method. For text elements, we collect text colors and cluster them into a palette. Each palette is up to 5 colors in this work. We get 18,768 / 2,315 / 2,278 valid data as train, validation, and test datasets. All design templates in the figures of this paper are from the Crello test dataset.

\subsection{Representation learning with masked color model}

\begin{figure*}[h]
\begin{center}
    \includegraphics[width=0.8\linewidth]{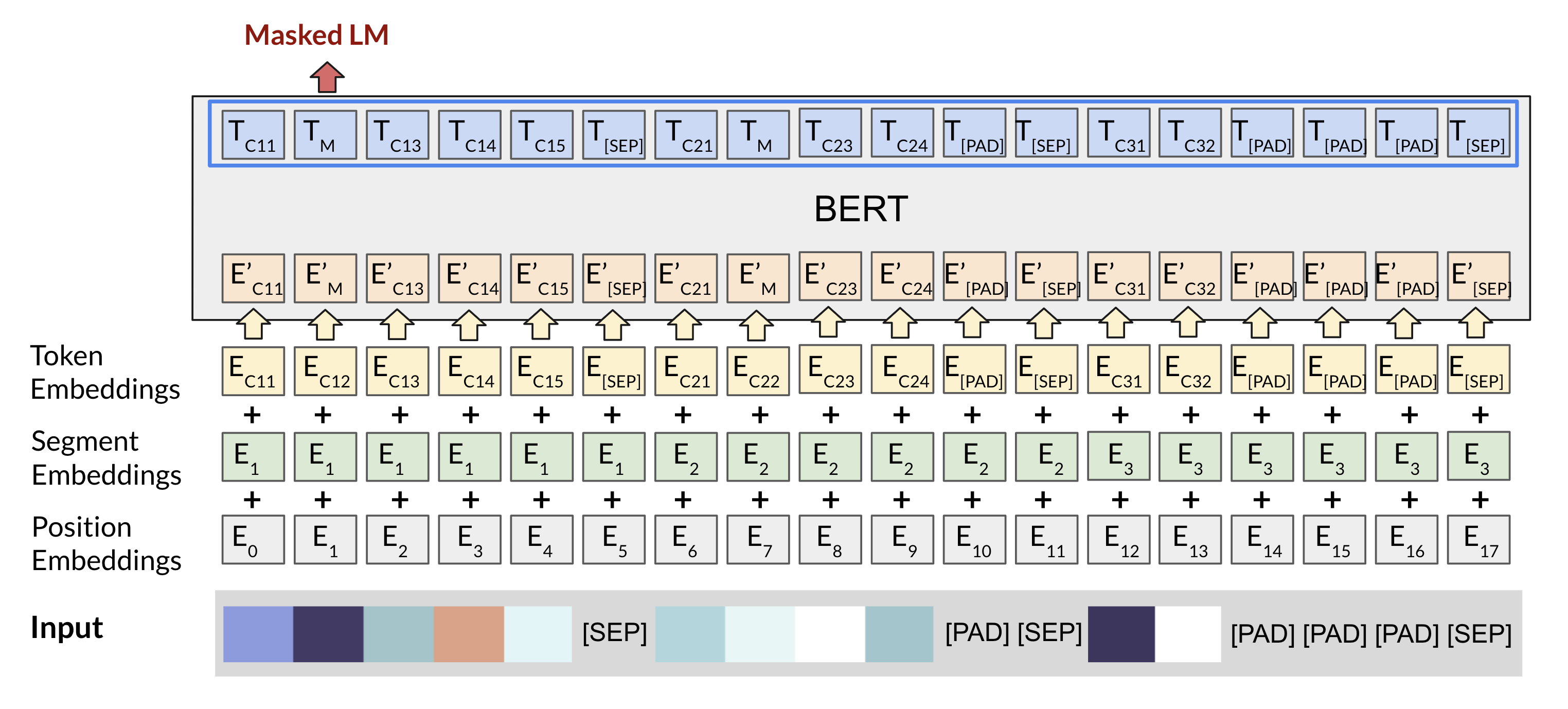}
\end{center}
  \caption{Masked color model for image-SVG-text color sequence.}
\label{fig:model}
\end{figure*}

We train the color embedding model in a similar way as the word embedding model. In natural language processing, the word embedding model is used to learn distributed representation, where the input is a text corpus and the output is a set of feature vectors that represent words. Similarly, in the color embedding model, a color signifies a word, a palette signifies a sentence, and multiple palettes in the same design signify a paragraph. 

For the input color corpus, we adopt CIELAB color space, which is more perceptually uniform than other color spaces \cite{kim2021dynamic}. The most general-used color space is the 24-bit RGB model. We convert RGB color data to CIELAB with a range of [0, 255], and assign each color to one of the bins in a $b \times b \times b$ histogram (we use $b=16$ in this work). For example, the color white(255, 255, 255) in RGB color space is labeled as the code $'15\_8\_8'$ in CIELAB color space with 16 bins. There are 796 color codes in the vocabulary of the train dataset. The color codes are converted into vectors and embedded in the space in the learning progress.

We obtain color embeddings by a masked color model based on pre-training BERT architecture \cite{kenton2019bert}. The masked color model in Figure~\ref{fig:model} is trained in the similar way as masked language model (Masked LM) in BERT. The model receives a fixed length of each palette as input. For a palette that is shorter than this fixed length, we will have to add the token [PAD] to the palette to make up the length. Another artificial token [SEP], is added to the end of a palette. For a given token, its input representation is constructed by summing the corresponding token, segment, and position embeddings. Here, $\{C_{1_{1}},\ldots C_{1_{5}}\}$ is for image color palette, $\{C_{2_{1}},\ldots C_{2_{4}}\}$ is for SVG color palette, $\{C_{3_{1}}, C_{3_{2}}\}$ is for text color palette. The palettes of image, SVG, and text, are respectively labeled with the segment number 1, 2, and 3. The segment embeddings are basically the palette number that is encoded into a vector. The trained model knows whether a particular color token belongs to a specific palette. The multi-palette representation can be achieved by segment embedings. 
% The color position in our current palette dataset is not as meaningful as the word position in sentences. We explore to remove the position embeddings in experimental validation section.

The masked color model randomly masks some percentage of the tokens from the input, and then predict the masked tokens based on its context. In our experiments, we mask $10\%$ of the tokens in each sequence at random and replace the chosen token with the [MASK] token $80\%$ of the time. Then, we use the standard cross entropy loss to optimize the pre-training task. 
The final embeddings from the Transformer self-attention mechanism can be used in downstream tasks, such as aesthetic rating prediction. 
In current work, we recommend colors in multi-palette by predicting masked colors with high probability.

\subsection{Color recommendation system}

In the existing creative applications for graphic documents, users are allowed to choose and edit a design template. However, when same visual elements are replaced, users may struggle with coordinating the colors in the design. To reduce user's work, we propose a system to recommend the specified colors and recolor the elements with the recommended colors automatically.

\begin{figure*}
\begin{center}
    \includegraphics[width=1.0\linewidth]{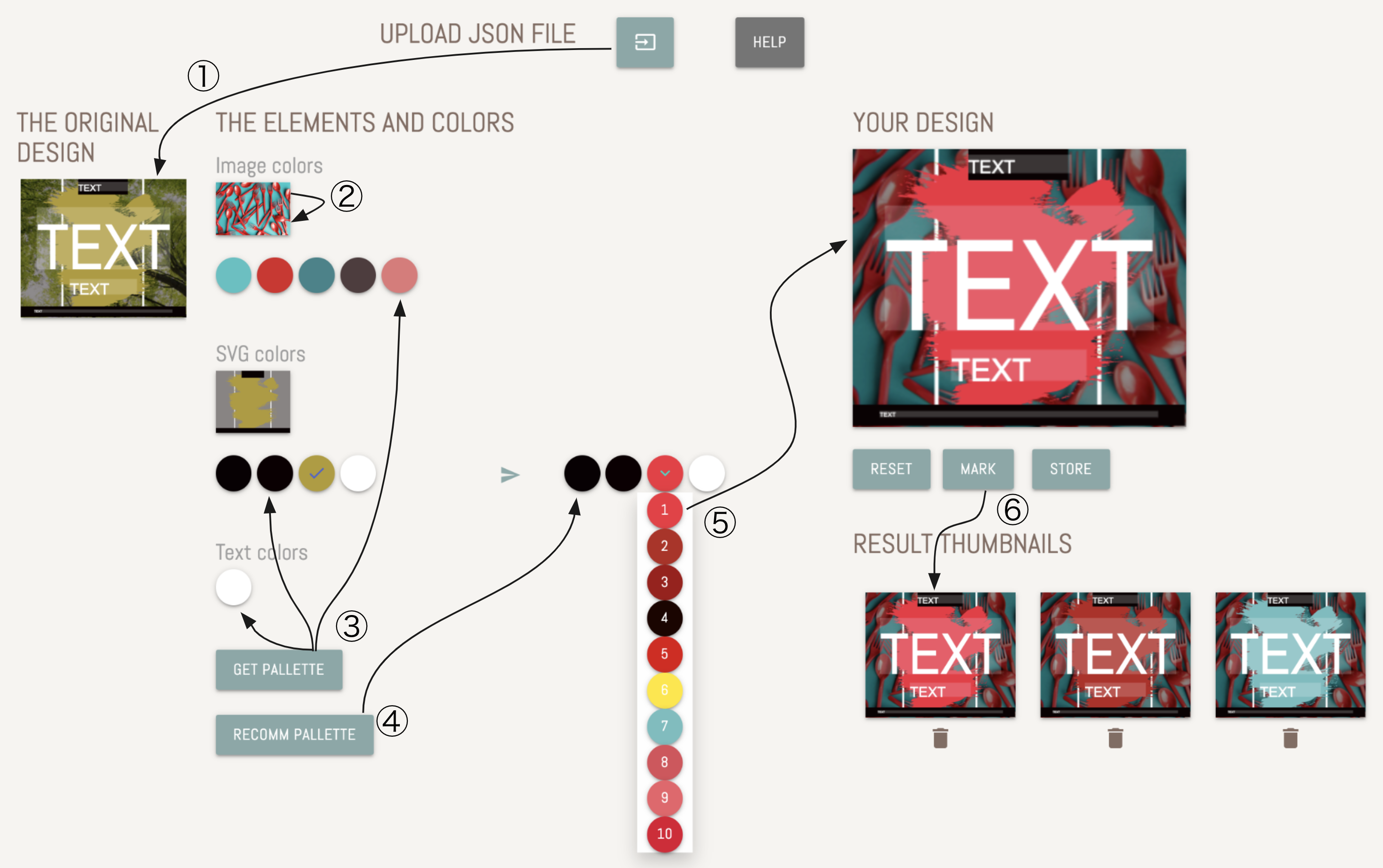}
\end{center}
   \caption{Interactive interface of color recommendation system for vector graphic documents contains six main operations: \textcircled{\raisebox{-0.9pt}{1}} Input a JSON file. \textcircled{\raisebox{-0.9pt}{2}} Replace image elements. \textcircled{\raisebox{-0.9pt}{3}} Get image-SVG-text palettes and select the colors for recoloring. \textcircled{\raisebox{-0.9pt}{4}} Get the recommended colors from the color recommendation engine. \textcircled{\raisebox{-0.9pt}{5}} Choose a recommended color and check the recolored result. \textcircled{\raisebox{-0.9pt}{6}} Mark the preferred results.}
\label{fig:system}
\end{figure*}

We create color recommendation engine with masked color model and develop an interactive user interface that enables users to obtain the coordinated colors for design. The color recommendation system supports basic selecting, interactive recommendation, and previewing functions shown in Figure~\ref{fig:system}. The design template is converted to a JSON file as the system input that contains complete element-specific configuration. The system parses a JSON object and reconstructs the design with separated visual elements. The system allows users to change the image elements and extracts the color palette of each element group. Users can select the colors for recoloring and then check the design results with the recommended colors. For SVG recoloring, the original color is changed to the recommended color by a simple interpolation method. For image recoloring, we use a palette-based photo recoloring method by k-means clustering \cite{chang2015palette} in this system.

\section{Experimental validation}

To evaluate the performance of our proposed approach, we compare with related work and a baseline model by both quantitative and qualitative evaluations. We adapted a Word2Vec-based model which is used in the related work \cite{kim2022colorbo}. The input for this model is the color token without segment embeddings. We also trained a BERT-based model without segment embeddings as a baseline to show the effectiveness of the segment embeddings for multi-palette representation.

\subsection{Quantitative evaluation}

We use 2278 color sequences of our test dataset in the quantitative experiments. We mask a color in color sequence randomly and evaluate the accuracy of the predicted color. We use top N accuracy that the true color is equal to any of the N most probable colors predicted by each model. The human eye sometimes cannot fully perceive subtle differences in color. In addition to accuracy, we also use visual similarity to measure the recommended colors. For similarity measurement, we calculate the distance between two colors using CIEDE2000 rather than Euclidean distance, which has exhibited good performance in predicting visual similarity between color palettes \cite{yang2020predicting, kim2021dynamic}. 

Firstly, we train 20 times and get the mean value of accuracy. The results of our model with and without segment and position embeddings are shown in Table~\ref{tab:quantitative_bert}. It is found that there is no difference between the models with and without position embeddings in the current dataset. Thus we pick up the best models trained by our method without position embeddings for following comparisons.

\begin{table}[b]
\begin{center}
\begin{tabular}{|l|l|c|c|c|c|}
\hline
\multicolumn{2}{|c|}{Embeddings} & \multicolumn{4}{|c|}{Accuracy$\uparrow$}\\
segment & position & @1 & @3 & @5 & @10 \\
\hline\hline
w/ & w/ & 0.27 & 0.45 & 0.53 & 0.63 \\
w/ & w/o & 0.27 & 0.44 & 0.52 & 0.62 \\
w/o & w/o & 0.16 & 0.30 & 0.38 & 0.50 \\
\hline
\end{tabular}
\end{center}
\caption{Quantitative comparison of our model with and without segment and position embeddings on top N accuracy (N = 1, 3, 5, 10). Here is the mean value of 20 trained models.}
\label{tab:quantitative_bert}
\end{table}

\begin{figure}[h]
\begin{center}
   \includegraphics[width=1.0\linewidth]{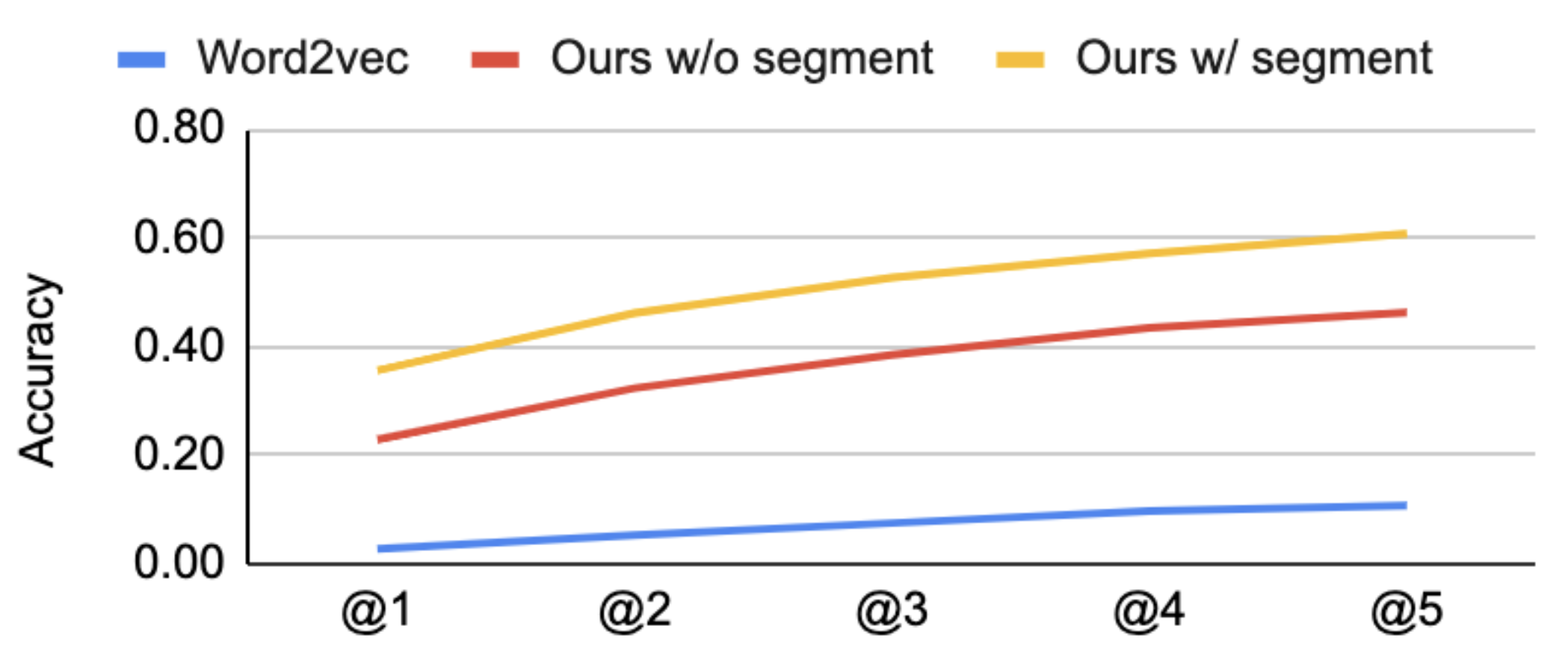}
\end{center}
   \caption{Quantitative comparison of our models with and without segment embeddings and the Word2Vec-based model on top N accuracy (N = 1, 2, 3, 4, 5).}
\label{fig:quantitative_accuracy}
\end{figure}

\begin{figure}[h]
\begin{center}
   \includegraphics[width=1.0\linewidth]{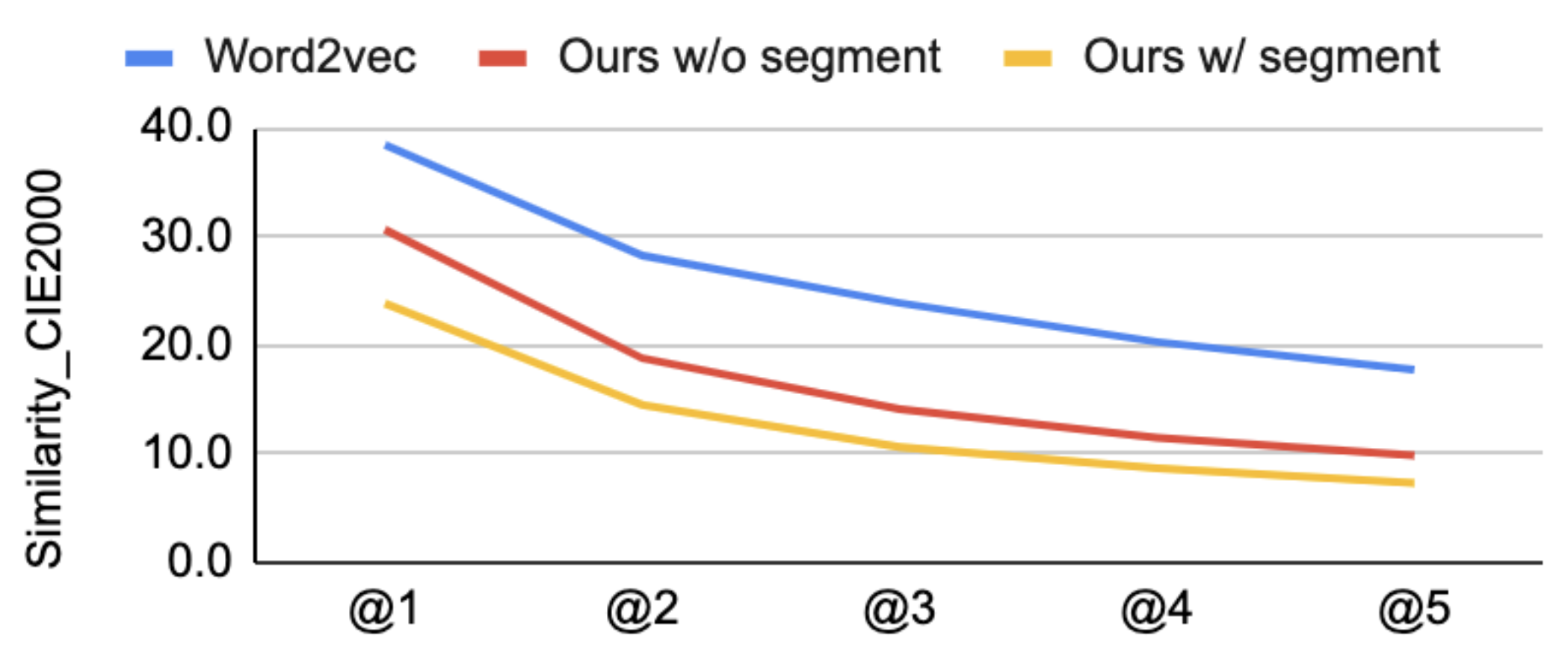}
\end{center}
   \caption{Quantitative comparison of our models with and without segment embeddings and the Word2Vec-based model on top N similarity (N = 1, 2, 3, 4, 5).}
\label{fig:quantitative_similarity}
\end{figure}

To compare our method with the Word2Vec-based model and the baseline model, we evaluate the accuracy and similarity of color prediction results by these three models.
The comparison results of accuracy are shown in Figure~\ref{fig:quantitative_accuracy}, Table~\ref{tab:quantitative_accuracy}, and the comparison results of similarity are shown in Figure~\ref{fig:quantitative_similarity}, Table~\ref{tab:quantitative_similarity}. Our method with segment embeddings provides significantly better results than the Word2Vec-based method and the baseline model. The results show that the segmentation is effective in multi-palette representation learning and brings better performance to color recommendation. 
We suggest providing more than two color candidates in recommendation applications with a high accuracy and users would like to find a desired color in top N recommended colors.

\begin{table}[h]
\begin{center}
\begin{tabular}{|l|c|c|c|c|c|}
\hline
& \multicolumn{5}{|c|}{Accuracy$\uparrow$}\\
Models & @1 & @2 & @3 & @4 & @5 \\
\hline\hline
Word2Vec & 0.03 & 0.05 & 0.08 & 0.10 & 0.11 \\
Ours w/o segment & 0.23 & 0.32 & 0.39 & 0.43 & 0.46 \\
\setrow{\bfseries} Ours w/ segment & 0.36 & 0.46 & 0.52 & 0.57 & 0.61 \\
\hline
\end{tabular}
\end{center}
\caption{Quantitative comparison of our models with and without segment embeddings and the Word2Vec-based model on top N accuracy (N = 1, 2, 3, 4, 5).}
\label{tab:quantitative_accuracy}
\end{table}

\begin{table}[h]
\begin{center}
\begin{tabular}{|l|c|c|c|c|c|}
\hline
& \multicolumn{5}{|c|}{Similarity$\downarrow$}\\
Models & @1 & @2 & @3 & @4 & @5 \\
\hline\hline
Word2Vec & 38.4 & 28.3 & 23.9 & 20.3 & 17.8 \\
Ours w/o segment & 30.6 & 18.8 & 14.1 & 11.5 & 9.9 \\
\setrow{\bfseries} Ours w/ segment & 23.8 & 14.5 & 10.7 & 8.7 & 7.4 \\
\hline
\end{tabular}
\end{center}
\caption{Quantitative comparison of our models with and without segment embeddings and the Word2Vec-based model on top N similarity (N = 1, 2, 3, 4, 5).}
\label{tab:quantitative_similarity}
\end{table}

\subsection{Qualitative evaluation}

\begin{figure}[h]
\begin{center}
   \includegraphics[width=1.0\linewidth]{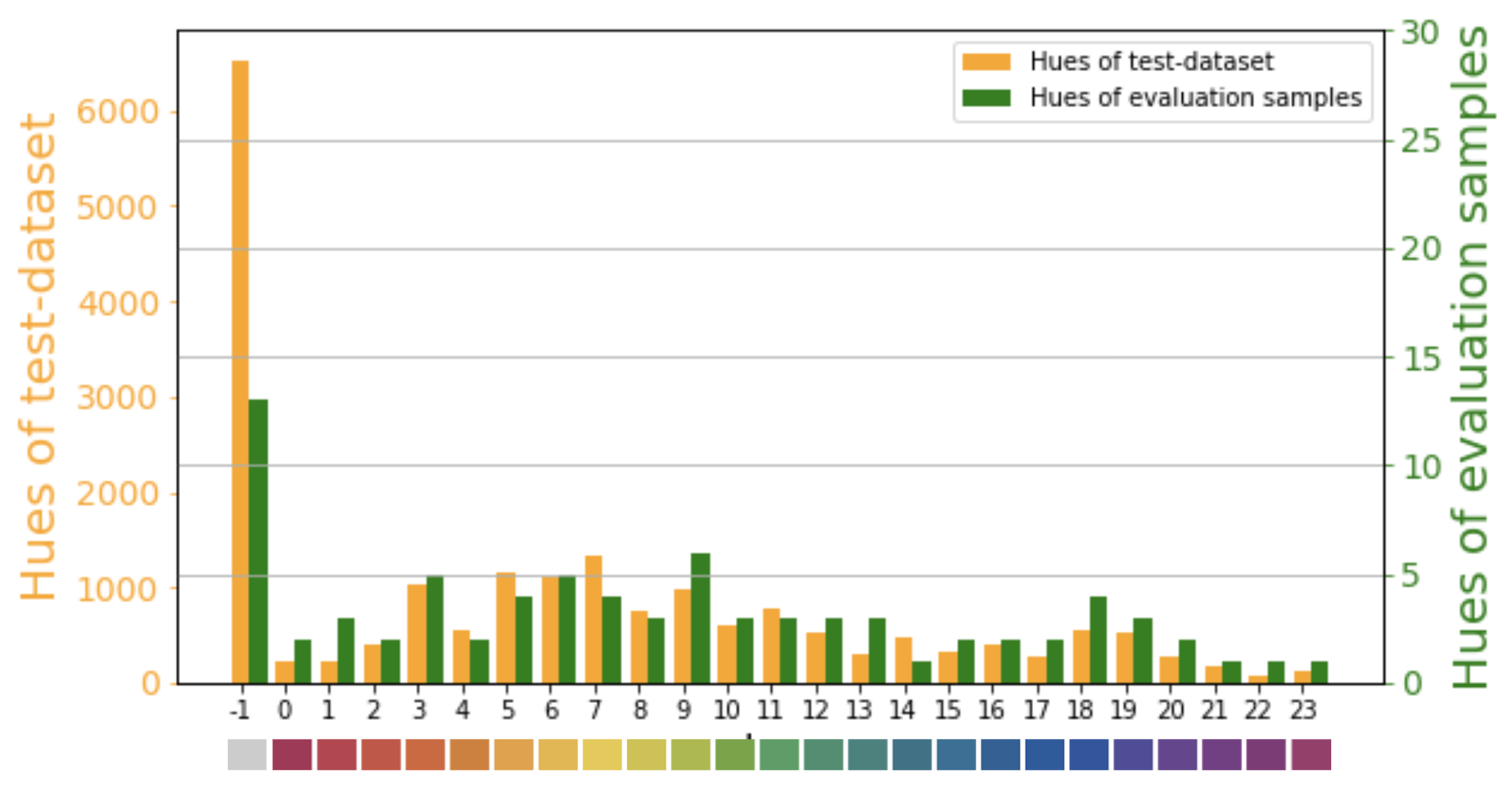}
\end{center}
   \caption{Hue distribution of selected colors in the evaluation samples and all colors in the Crello test dataset. Hue orders are based on the Practical Color Co-ordinate System, and we denote the neutral color as -1.}
\label{fig:eval_hues}
\end{figure}

\begin{figure}[h]
\begin{center}
   \includegraphics[width=1.0\linewidth]{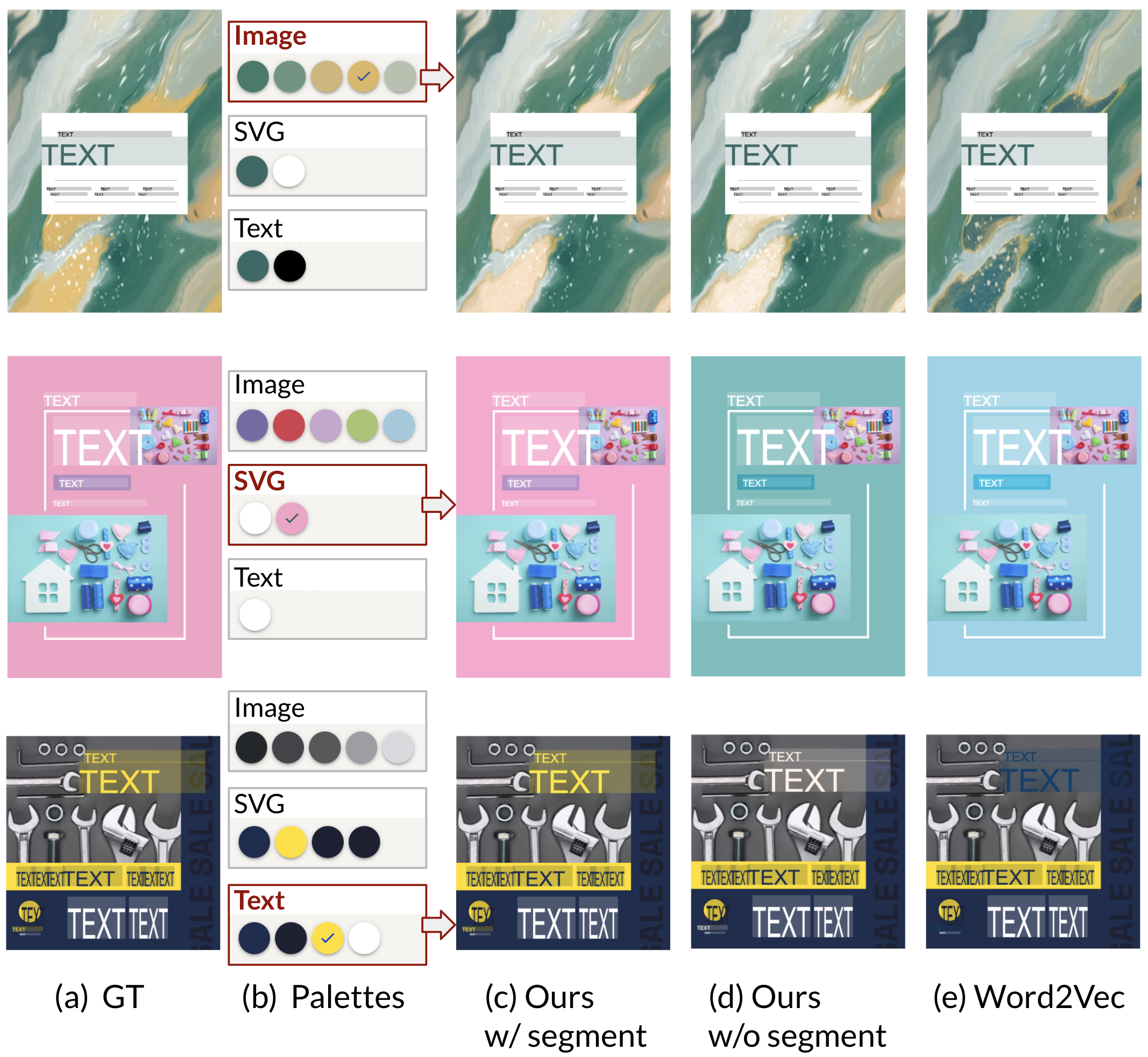}
\end{center}
   \caption{Color recommendation results with our proposed BERT-based models with and without segment embeddings, and the Word2Vec-based model. The three samples are recolored with one image color, one SVG color, and one text color respectively.}
\label{fig:recomm_results}
\end{figure}

Considering that color performance depends on human perception, 
% it is essential to provide a quantitative evaluation by human observers. 
we conducted qualitative evaluation to verify the performance of the recommended results. We randomly selected 80 templates from the Crello test dataset and randomly selected one color for recoloring in multi-palette, that can be in image, or SVG, or text element. The hue distribution of selected color in the evaluation samples and all colors in the test dataset are shown in Figure~\ref{fig:eval_hues}. The neutral colors in some 
visually imperceptible elements are excluded during random selection, e.g. a text color with very small font size is ignored in this experiment. The recolored designs by the top1 recommended colors of each model are shown in Figure~\ref{fig:recomm_results}. 

We pick up one original design (GT), and the top2 recommended results from three models including our model with segment, the baseline model without segment, and the Word2Vec-based model. These seven designs are arranged together in an evaluation question. The participants are asked to select at most three good designs and three bad designs from seven designs. We recruited 84 participants with 68 non-designers and 16 graphic designers.

\begin{figure}[h]
\begin{center}
   \includegraphics[width=1.0\linewidth]{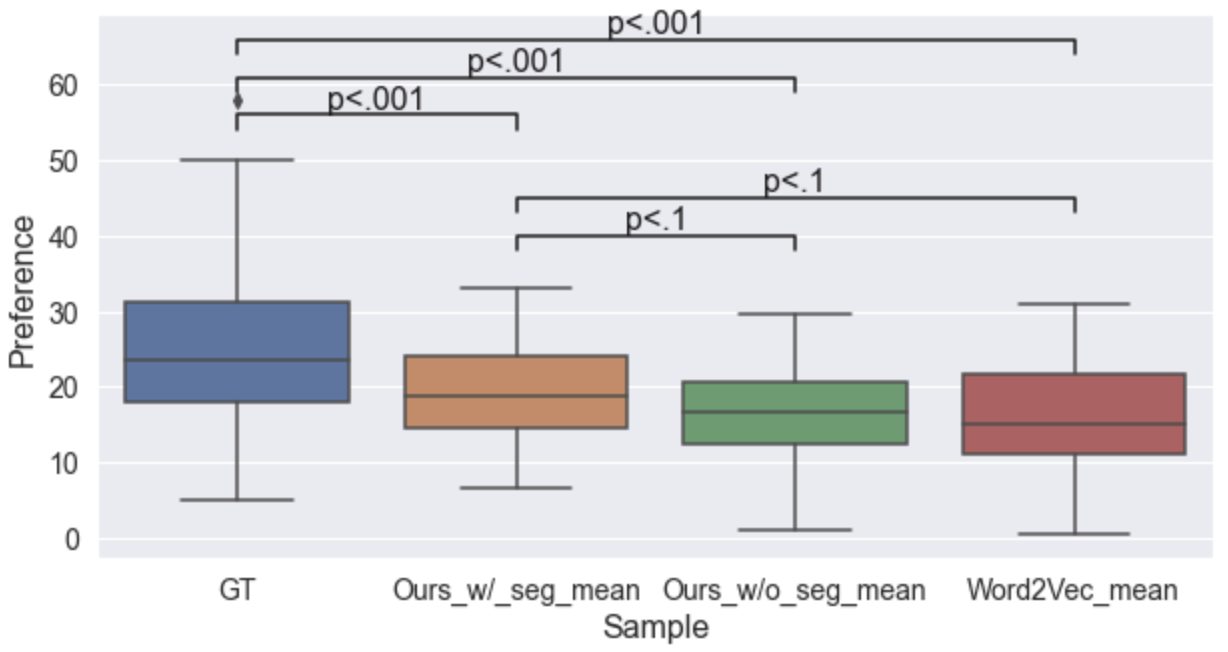}
\end{center}
   \caption{Evaluation results of good design from non-designers.}
\label{fig:eval_nondesigner_g}
\end{figure}

\begin{figure}[h]
\begin{center}
   \includegraphics[width=1.0\linewidth]{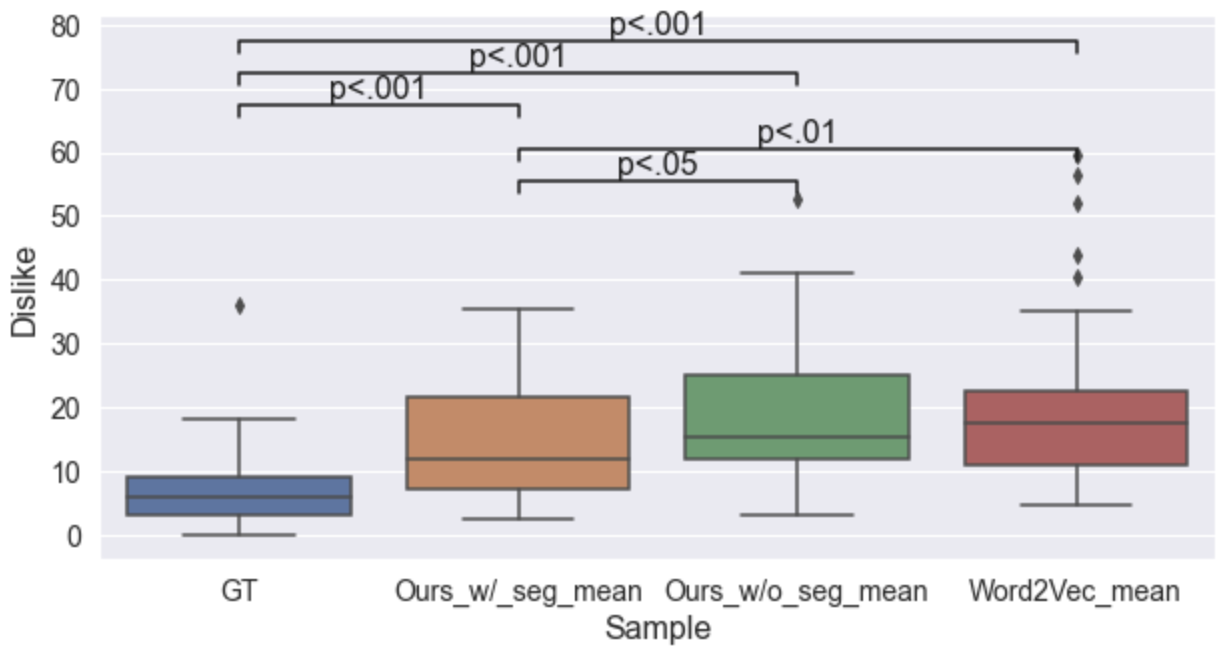}
\end{center}
   \caption{Evaluation results of bad design from non-designers.}
\label{fig:eval_nondesigner_ng}
\end{figure}

For non-designers, the evaluation results of good and bad design selections are shown in Figure~\ref{fig:eval_nondesigner_g} and Figure~\ref{fig:eval_nondesigner_ng}. Here is the mean value of top2 recommendation results by three models. We can find that though the results by our proposed model with segment embeddings get worse performance than GT, they have higher preference and lower dislike than the baseline model without segment and the Word2Vec-based model ($p<0.1$). There is no significant difference between the baseline model and the Word2Vec-based model.

\begin{figure}[h]
\begin{center}
   \includegraphics[width=1.0\linewidth]{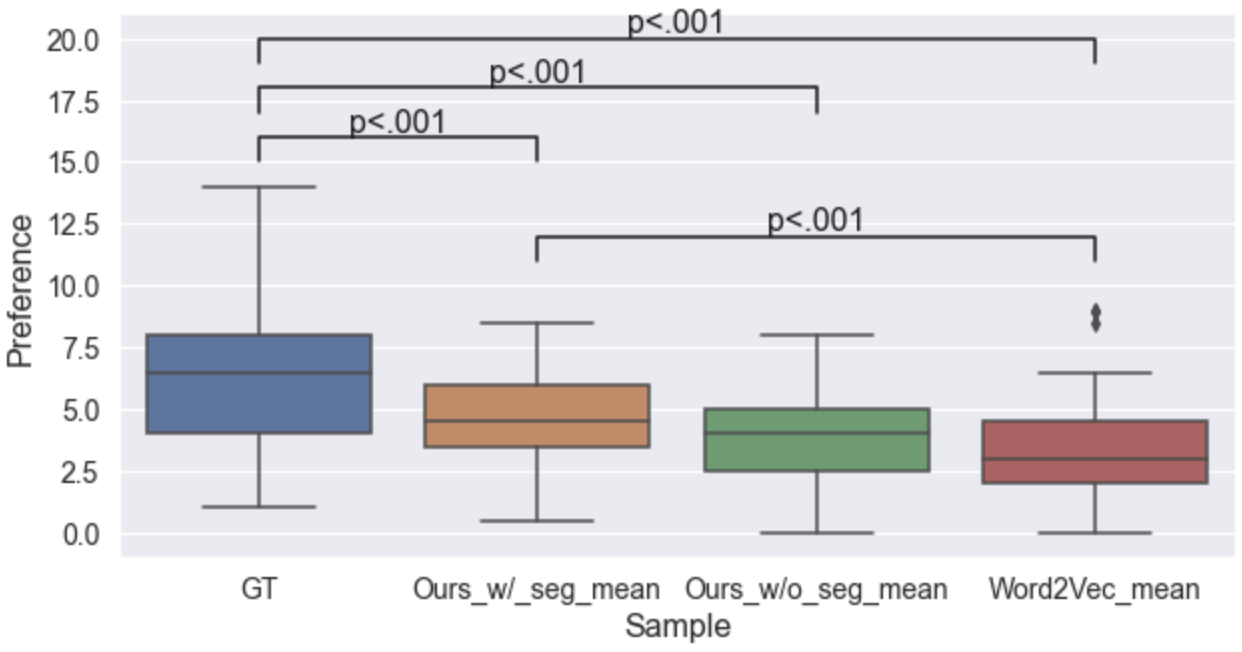}
\end{center}
   \caption{Evaluation results of good design from designers.}
\label{fig:eval_designer_g}
\end{figure}

\begin{figure}[h]
\begin{center}
   \includegraphics[width=1.0\linewidth]{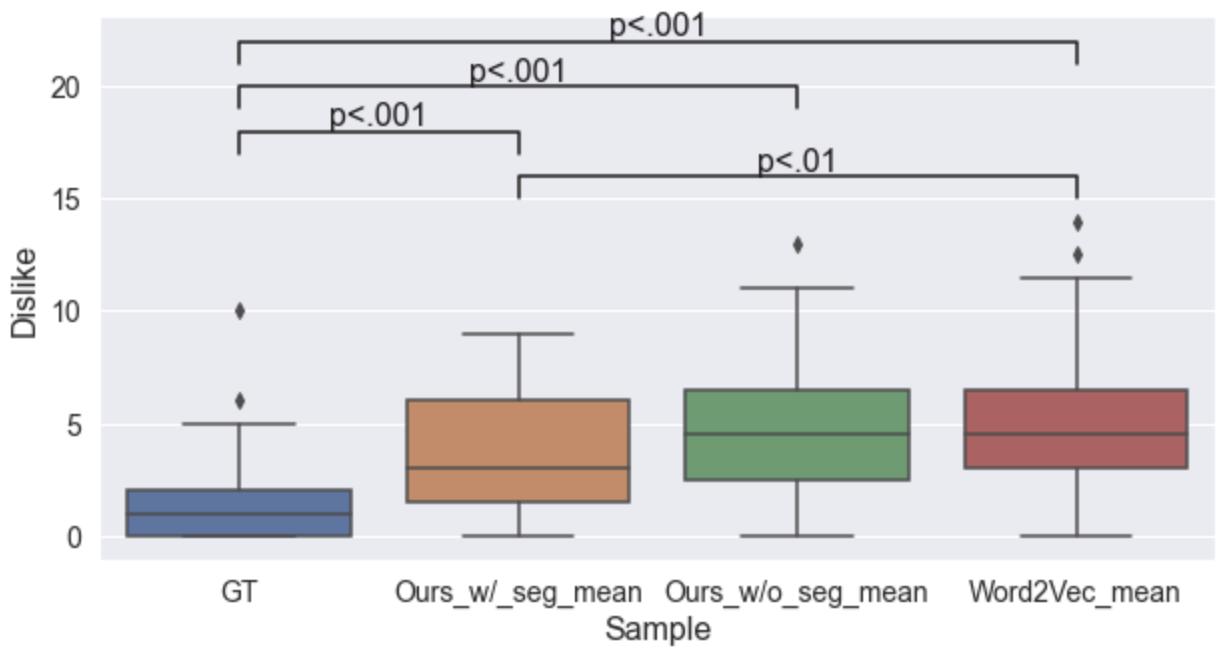}
\end{center}
   \caption{Evaluation results of bad design from designers.}
\label{fig:eval_designer_ng}
\end{figure}

 For designers, the evaluation results of good and bad design selections are shown in Figure~\ref{fig:eval_designer_g} and Figure~\ref{fig:eval_designer_ng}. The results are similar with the results from non-designers. Moreover, designers evaluate that our model with segment has significantly better performance than the Word2Vec model (preference: $p<0.001$, dislike: $p<0.01$). From the evaluation results of non-designers and designers, the most obvious finding is that our proposed model has better performance than the Word2Vec-based model. Moreover, the segment embedding is necessary to multi-paltte representation. 
%  On the other hand, the design templates in Crello dataset are proved to have appropriate color palettes. This dataset is suitable for color palette representation learning.

\subsection{Interview study}

\begin{figure}[b]
\begin{center}
   \includegraphics[width=1.0\linewidth]{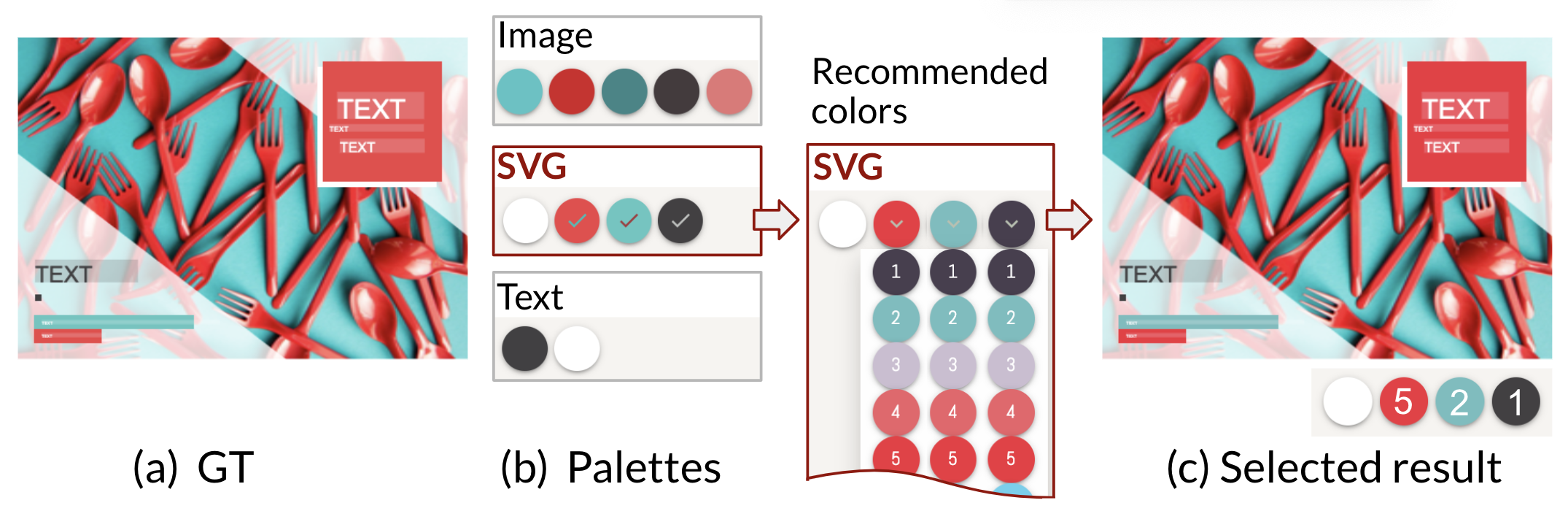}
\end{center}
   \caption{Color recommendation results for recoloring three colors in SVG elements. The number shown on the palette color is the recommended ranking.}
\label{fig:3color_recomm}
\end{figure}

To access the color recommendation system as in Figure~\ref{fig:system}, we collected qualitative feedback from 12 professional designers aged 20-39. We provided a short tutorial of our system and asked the participants to explore color recommendation for recoloring one color in visual elements. For the elements in some templates, there are more than one dominant colors. We also asked participants to explore the recommendation of more than one color. Here is a sample of recoloring three colors in SVG elements in Figure~\ref{fig:3color_recomm}. We prepared design templates and some image samples from the Crello test dataset. 

\twocolumn[{
\begin{center}
    \captionsetup{type=table}
    \begin{tabular}{|l|l|}
    \hline
    \textbf{Q1} Struggle with choosing colors in creative design works&\includegraphics[scale=0.25]{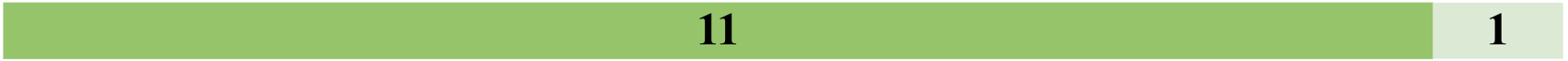} \\
    \textbf{Q2} Color recommendation tool is useful&\includegraphics[scale=0.25]{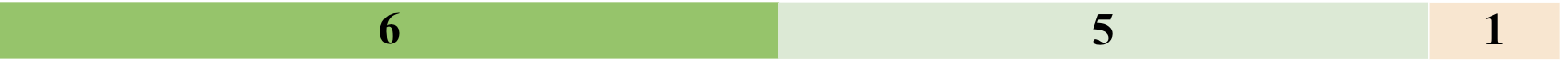} \\
    \textbf{Q3} This recommendation system is easy to use&\includegraphics[scale=0.25]{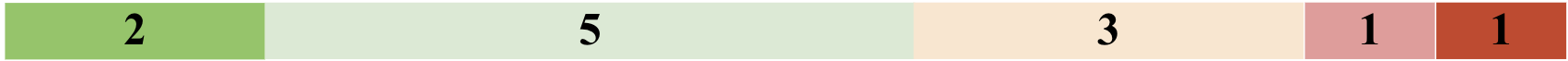} \\
    \textbf{Q4} Recommended results for one color look good&\includegraphics[scale=0.25]{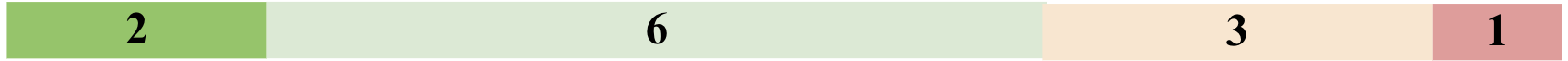} \\
    \textbf{Q5} Recommended results for more than one color look good&\includegraphics[scale=0.25]{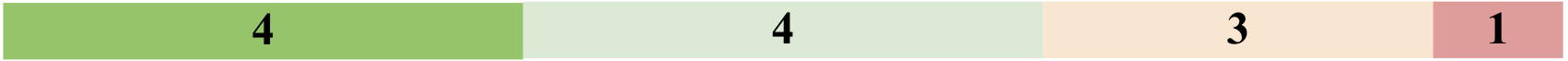} \\
    \textbf{Q6} Like to use this recommendation system for work&\includegraphics[scale=0.25]{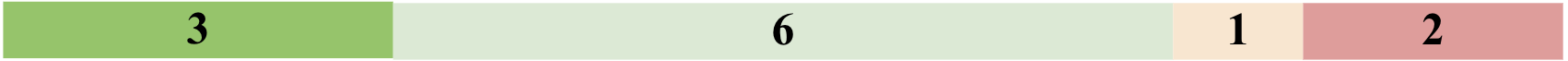} \\
    & \includegraphics[scale=0.25]{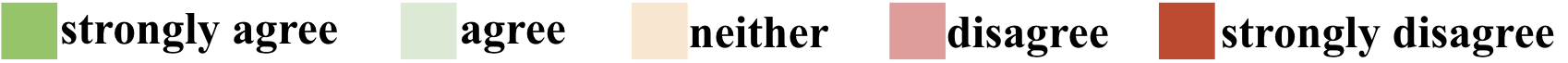}\\
    \hline
    \end{tabular}
    \captionof{table}{Evaluation results from 12 designers in the interview study.}
    \label{tab:interview_study}
\end{center}
}]

We asked participants to replace the image elements in design templates with image samples. The participants could also upload their own images. The participants filled out a questionnaire after some trials. There are six questions and each question has five choices, as strongly agree, agree, neither, disagree, and strongly disagree. 

The results are shown in Table~\ref{tab:interview_study}. We observed the high demand for color recommendation systems in creative design works as in Q1 and $91.7\%$ participants answered that a color recommendation tool is useful for graphic design work in Q2. For Q3 and Q6, $58.3\%$ participants answered that our current system is easy to use, and $75\%$ participants liked to use our recommendation system for work. For the recommendation results, $66.7\%$ participants can get a satisfied recommended colors by our system in one color and more than one color recoloring tasks as in Q4 and Q5. Generally, we received positive feedback for our color recommendation system and the recommended results from designers.

\subsection{Limitations}

\textbf{Accuracy decreases when masked color number increases.} We generate the color sequence with a maximum 15 colors and train our model with masking 10$\%$ of the token in each sequence. That is, only 1 color is masked for prediction in most sequences in the training process. When the masked color number increases, the prediction accuracy decreases significantly in Table~\ref{tab:accuracy_multi_mask}. 

\begin{table}[h]
\begin{center}
\begin{tabular}{|l|c|c|c|c|c|}
\hline
Masked colors & 1 & 2 & 3 & 4 & 5 \\\hline
Accuracy@1$\uparrow$ & 0.36 & 0.29 & 0.24 & 0.20 & 0.17 \\
\hline
\end{tabular}
\end{center}
\caption{Top 1 accuracy for predicting different numbers of masked colors.}
\label{tab:accuracy_multi_mask}
\end{table}

\textbf{Lack of diversity in recommended colors when more than one color is masked.} Though users can freely combine the recommended colors and designers can get a satisfactory design by our recommendation system in interview study, how to recommend a complete palette for each element by a learned model remains a problem. The recommended colors in the same element group are highly similar as in Figure~\ref{fig:3color_recomm}. Furthermore, neutral colors have a semantic value to a color palette and we do not filter out these high-frequency colors. Neutral colors have high frequency in the dataset, and it has higher probability to be recommended than chromatic colors. 
% For example, white and black tend to have high ranking than appropriate chromatic colors in top N recommendation.

\section{Conclusions}

We proposed a masked color model for multi-palette representation to recommend colors for vector graphic documents and developed an interactive system of recoloring the specified colors in visual elements. The performance of the proposed system is experimentally verified through both quantitative and qualitative evaluations compared to the state-of-the-art method of a Word2Vec-based model and the baseline model. To our knowledge, our method opens the door to recommend colors for vector graphic design based on multi-palette of visual elements. We will explore to improve the performance of a complete palette recommendation and fine-turn our model for the practical applications in the future work.

{\small
\bibliographystyle{ieee_fullname}
\bibliography{egbib}
}

\end{document}